\documentclass[runningheads]{llncs}
\usepackage{graphicx}

\usepackage{amssymb} 
\usepackage{amsfonts}
\usepackage{amsmath}
\usepackage{bm}

\usepackage{subcaption}
\usepackage{multirow}
\usepackage{diagbox}
\usepackage{booktabs}

\usepackage{hyperref}

\usepackage[dvipsnames]{xcolor}
\usepackage{pgfplots}
\pgfplotsset{compat=1.17}
\usetikzlibrary{calc}

\title{On the dice loss gradient\\and the ways to mimic it}
\titlerunning{On the dice loss gradient and the ways to mimic it}

\author{Hoel \textsc{Kervadec}\textsuperscript{1}\\
Marleen \textsc{de Bruijne}\textsuperscript{1,2}}

\authorrunning{\textsc{Kervadec} et \textsc{de Bruijne}}
\institute{1: Erasmus MC, Rotterdam, The Netherlands\\
2: University of Copenhagen, Denmark}

\begin{document}
        \maketitle
        \begin{abstract}
                In the past few years, in the context of fully-supervised semantic segmentation, several losses---such as cross-entropy and dice---have emerged as \emph{de facto} standards to supervise neural networks.
                The Dice loss is an interesting case, as it comes from the relaxation of the popular Dice coefficient; one of the main evaluation metric in medical imaging applications.
                In this paper, we first study theoretically the gradient of the dice loss, showing that concretely it is a weighted negative of the ground truth, with a very small dynamic range.
                This enables us, in the second part of this paper, to mimic the supervision of the dice loss, through a simple element-wise multiplication of the network output with a negative of the ground truth.
                This rather surprising result sheds light on the practical supervision performed by the dice loss during gradient descent.
                This can help the practitioner to understand and interpret results while guiding researchers when designing new losses.

        \end{abstract}

        \section{Introduction}
                Loss functions play a crucial role when training a deep neural networks---as they act as the interface between the labels and network predictions---, and some of them are now seen as \emph{de facto} standards for image segmentation, such as the cross-entropy and dice losses \cite{isensee2021nnu,eelbode2020optimization,ma2021loss}.
                One clear example is the default configuration of nn-UNet \cite{isensee2021nnu}, a self-configuring U-Net architecture that consistently ranks high in medical challenges.
                There, the choice of cross-entropy plus dice is seen as optimal, and is a fixed parameter of the framework.

                As far as we are aware, while several works proposed modifications of the standard dice loss definition to better handle imbalanced problems (V-Net \cite{milletari2016v} and the GDL \cite{sudre2017generalised}), less works are studying it from a theoretical point of view.
                In \cite{eelbode2020optimization}, the authors show that Dice and Jaccard are absolute and relative approximate of each others, and show that metric sensitive losses are superior in context where evaluation is performed with Dice Score \cite{maier2022metrics}.
                In the context of binary segmentation, \cite{nordstrom2020calibrated} show a quasi-concave lower-bound of the Dice loss that might be more stable during training.

                We complement earlier analyses by studying the behavior of the dice loss \emph{from a gradient point of view} during back-propagation.
                We show that the gradient is effectively a weighted negative of the label, and can behave in counter-intuitive ways. 
                We also show that the magnitude of its gradient has a very small dynamic range, which in the second part of our paper we mimic in a simple but effective way.
                This, we believe, shed new insights on the behavior of dice loss, and segmentation losses for neural networks in general.

        \section{Analysis of the dice loss}
                \subsection{Definition}
                        The Sørensen--Dice coefficient is a very common way to measure the overlap between two areas \cite{maier2022metrics}---in our case a fully annotated ground truth $y$ and a predicted segmentation $s$---and can be formally written for class $k \in \mathcal K$ as:
                        \begin{equation}
                                \label{eq:dsc}
                                \text{DSC}(y, s;k) := \frac{2\left|\Omega_y^{(k)} \cap \Omega_s^{(k)}\right|}{\left|\Omega_y^{(k)}\right| + \left|\Omega_s^{(k)}\right|},
                        \end{equation}
                        with $\Omega \subset \mathbb R^D$ a $D$-dimensional image space, $y^{(\cdot,\cdot)}: (\Omega \times \mathcal K) \rightarrow \{0, 1\}$ representing the ground-truth as a binary function, and $s^{(\cdot,\cdot)}: (\Omega \times \mathcal K) \rightarrow \{0, 1\} $ the predicted segmentation. 
                        $\mathcal K = \{0: \texttt{background}, 1: \texttt{class1}, ..., K: \texttt{classK} \}$ is the set of classes to segment, which can be binary ($K=1$) or multi-class ($K \geq 2$). Note that here the background is represented explicitly at index $0$, so that $|\mathcal K| = K + 1$. A summary of all notation is available in the supplementary material \ref{sec:notation}.

                        $y^{(i,k)} = 1$ means that voxel $i \in \Omega$ belongs to class $k$  and $y^{(i,k)} = 0$ means that it does not.
                        We denote $\Omega_y^{(k)} := \{i \in \Omega | y^{(i, k)}=1\} \subseteq \Omega$ the subset of the image space where $y$ is of class $k$.
                        The same follows for $\Omega_s^{(k)}$. 

                        Representing each voxel as a binary variable, it can be rewritten:
                        \begin{equation}
                                \text{DSC}(y, s;k) = \frac{2\sum_{i \in \Omega}y^{(i,k)}s^{(i,k)}}{\sum_{i \in \Omega}\left[y^{(i,k)} + s^{(i,k)}\right]}.
                        \end{equation}

                        From there, it becomes trivial to relax its definition to continuous probabilities (with $s_{\boldsymbol\theta}^{(i,k)} \in [0, 1]$ the neural network predicted probabilities; $\boldsymbol\theta$ being the network parameters) and write it as a loss to be minimized:
                        \begin{equation}
                                \label{eq:l_dsc}
                                \mathcal L_\text{DSC}(y, s_{\boldsymbol\theta}) := \frac{1}{|\mathcal K|} \sum_{k \in \mathcal K} \left(1 - \frac{2\sum_{i \in \Omega}y^{(i,k)}s_{\boldsymbol\theta}^{(i,k)}}{\sum_{i \in \Omega}\left[y^{(i,k)} + s_{\boldsymbol\theta}^{(i,k)}\right]}\right).
                        \end{equation}

                        This loss, or some of its variants \cite{milletari2016v,sudre2017generalised,ma2021loss,eelbode2020optimization}, is often used in medical image segmentation, either as stand-alone or in combination ($\lambda_1 \mathcal L_1 + \lambda_2 \mathcal L_2$, with $\lambda_1,\lambda_2 \in \mathbb R$) with the cross-entropy loss:
                        \begin{equation}
                                \label{eq:l_ce}
                                \mathcal L_\text{CE}(y, s_{\boldsymbol\theta}) := \frac{1}{|\mathcal K||\Omega|} \sum_{k \in \mathcal K} \sum_{i \in \Omega} -y^{(i,k)}\log\left(s_{\boldsymbol\theta}^{(i,k)}\right) .
                        \end{equation}
                \subsection{Gradients}
                        \subsubsection{Derivation}
                                Taking a look at the gradients of the dice loss, we have:
                                \begin{align}
                                        \frac{\partial \mathcal L_\text{DSC}}{\partial s_{\boldsymbol\theta}^{(i,k)}} &= \frac{\partial\left(- \frac{2\sum_{i \in \Omega}y^{(i,k)}s_{\boldsymbol\theta}^{(i,k)}}{\sum_{i \in \Omega}\left[y^{(i,k)} + s_{\boldsymbol\theta}^{(i,k)}\right]}\right)}{\partial s_{\boldsymbol\theta}^{(i,k)}} = \frac{\partial \left( -\frac{2I^{(k)}}{U^{(k)}} \right)}{\partial s_{\boldsymbol\theta}^{(i, k)}} \nonumber\\
                                        &= -2 \left(\frac{\frac{\partial I^{(k)}}{\partial s_{\boldsymbol\theta}^{(i,k)}}U^{(k)} - I^{(k)}\frac{\partial U^{(k)}}{\partial s_{\boldsymbol\theta}^{(i,k)}}}{\left(U^{(k)}\right)^2} \right),
                                        \label{eq:init_derivation}
                                \end{align}
                                with $I^{(k)} = \sum_{i \in \Omega}y^{(i,k)}s_{\boldsymbol\theta}^{(i,k)} $ and $U^{(k)} = \sum_{i \in \Omega}\left[y^{(i,k)} + s_{\boldsymbol\theta}^{(i,k)}\right]$.
                                Reminding that $\frac{\partial s_{\boldsymbol\theta}^{(j,k)}}{\partial s_{\boldsymbol\theta}^{(i,k)}} = 1$ if $i=j$ and $0$ otherwise, we have:  $\frac{\partial I^{(k)}}{\partial s_{\boldsymbol\theta}^{(i,k)}} = \sum_{j \in \Omega} \frac{\partial y^{(j,k)}s_{\boldsymbol\theta}^{(j,k)}}{\partial s_{\boldsymbol\theta}^{(i,k)}} = y^{(i,k)}$ and $\frac{\partial U^{(k)}}{\partial s_{\boldsymbol\theta}^{(i,k)}} = \sum_{j \in \Omega} \frac{\partial s_{\boldsymbol\theta}^{(j,k)}}{\partial s_{\boldsymbol\theta}^{(i,k)}} = 1$.
                                Hence:
                                \begin{align*}
                                        \frac{\partial \mathcal L_\text{DSC}}{\partial s_{\boldsymbol\theta}^{(i, k)}} = \frac{-2\left(y^{(i, k)}U^{(k)} - I^{(k)}\right)}{\left(U^{(k)}\right)^2} \\
                                        & & \square
                                \end{align*}
                        \subsubsection{A two-valued gradient}
                                One will quickly notice a peculiar situation:
                                \begin{equation}
                                        \label{eq:dynamic_weight}
                                        \frac{\partial \mathcal L_\text{DSC}}{\partial s_{\boldsymbol\theta}^{(i, k)}} = \begin{cases}
                                                \frac{-2\left(U^{(k)} - I^{(k)}\right)}{\left(U^{(k)}\right)^2} & \text{if } y^{(i, k)} = 1,  \\
                                                \frac{2I^{(k)}}{\left(U^{(k)}\right)^2} & \text{otherwise}.
                                        \end{cases}
                                \end{equation}

                                \begin{figure}[h]
                                        \centering
                                        \begin{subfigure}[t]{0.45\textwidth}
                                                \centering
                                                \includegraphics[height=80pt]{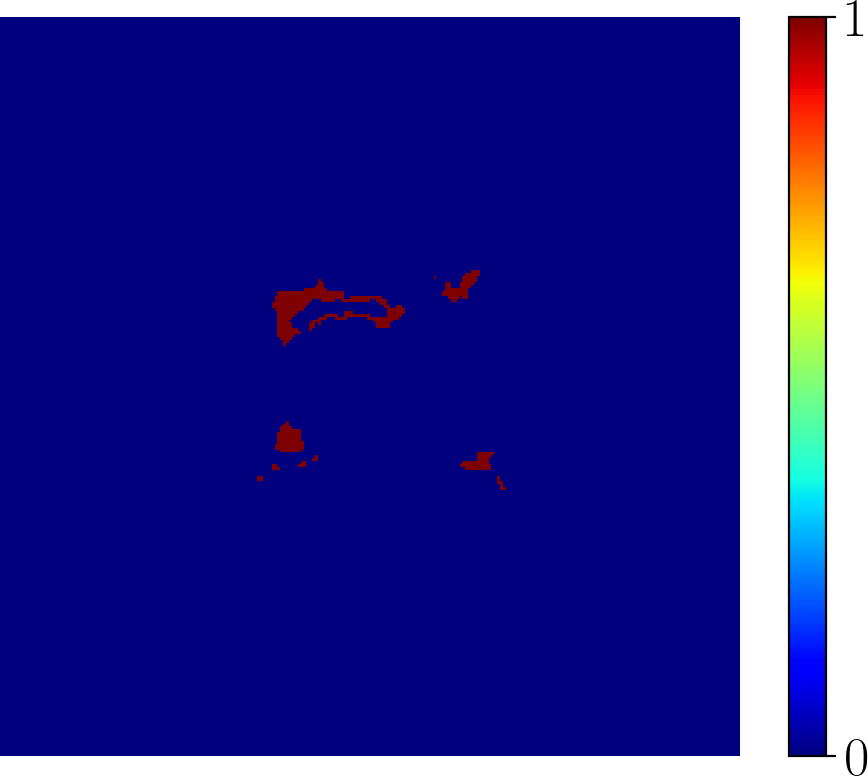}
                                                \caption{$y^{(1,k)}$}
                                                \label{subfig:labels}
                                        \end{subfigure}
                                        \begin{subfigure}[t]{0.45\textwidth}
                                                \centering
                                                \includegraphics[height=80pt]{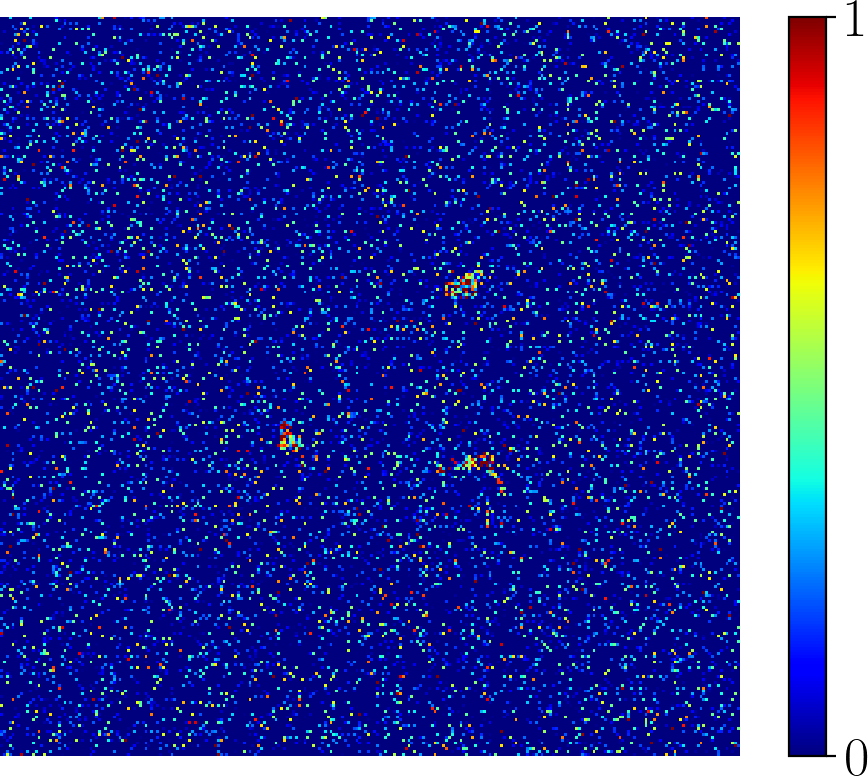}
                                                \caption{$s_{\boldsymbol\theta}^{(1, k)}$}
                                                \label{subfig:softmax}
                                        \end{subfigure}
                                        \\
                                        \begin{subfigure}[t]{0.45\textwidth}
                                                \centering
                                                \includegraphics[height=80pt]{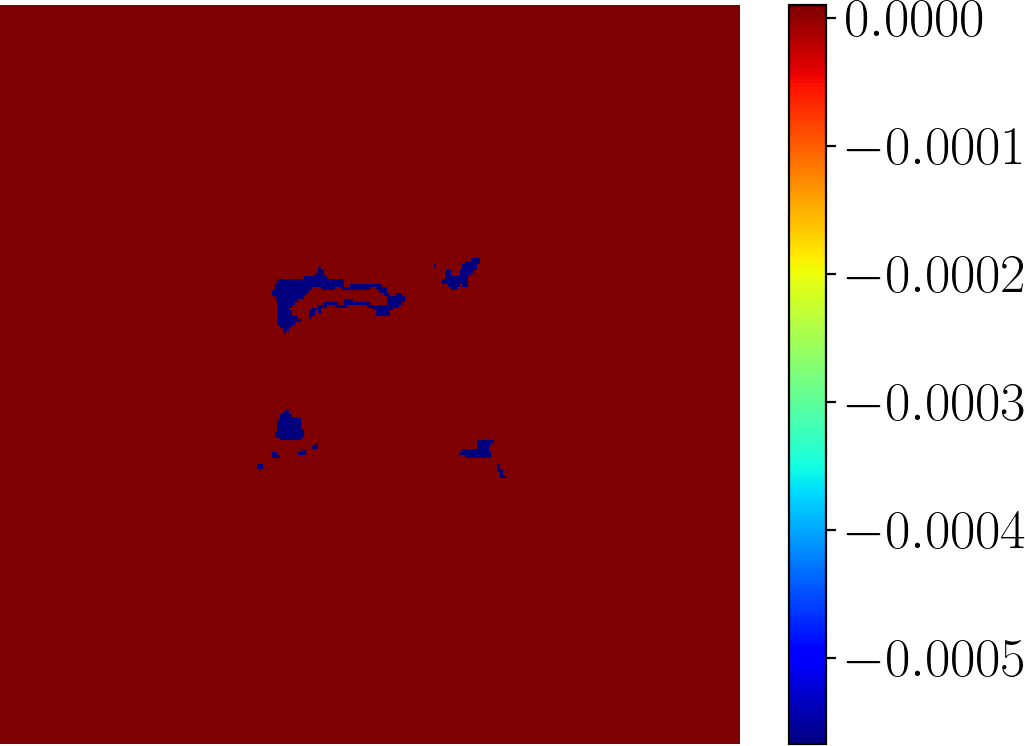}
                                                \caption{$\frac{\partial \mathcal L_\text{DSC}}{\partial s_{\boldsymbol\theta}^{(0, k)}} + \frac{\partial \mathcal L_\text{DSC}}{\partial s_{\boldsymbol\theta}^{(1, k)}}$}
                                                \label{subfig:dsc_grad}
                                        \end{subfigure}
                                        \begin{subfigure}[t]{0.45\textwidth}
                                                \centering
                                                \includegraphics[height=80pt]{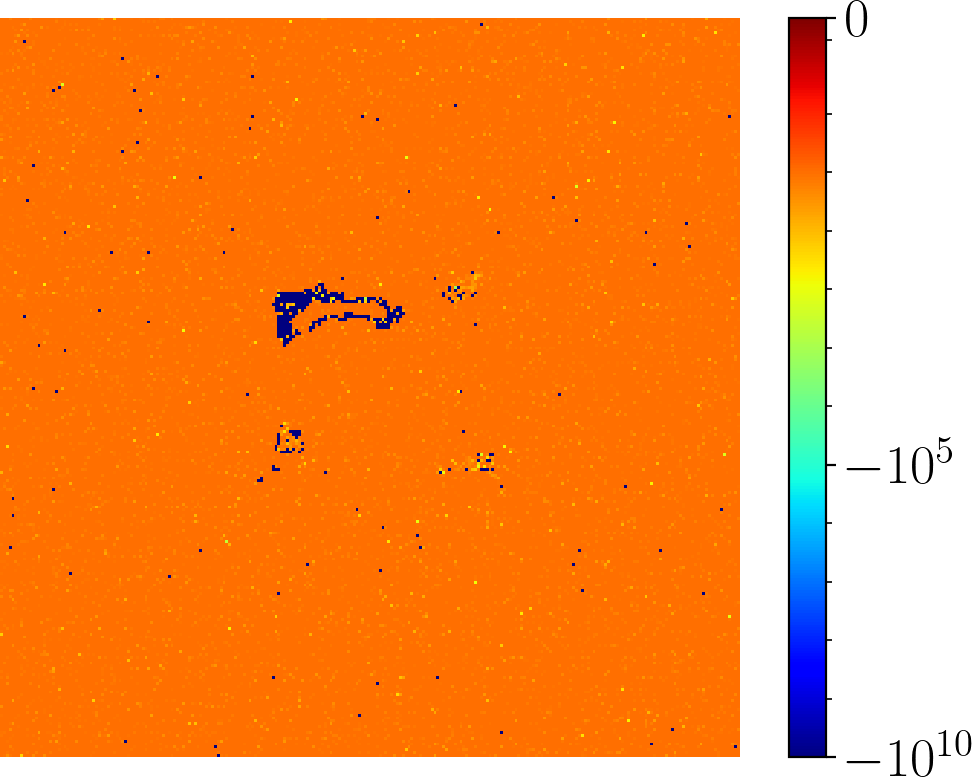}
                                                \caption{$\frac{\partial \mathcal L_\text{CE}}{\partial s_{\boldsymbol\theta}^{(0, k)}} + \frac{\partial \mathcal L_\text{CE}}{\partial s_{\boldsymbol\theta}^{(1, k)}}$}
                                                \label{subfig:ce_grad}
                                        \end{subfigure}
                                        \caption{Partial derivatives of $\mathcal L_\text{DSC}$ and $\mathcal L_\text{CE}$ wrt. ground truth $y$ and prediction $s_{\boldsymbol\theta}$. Notice the noise in (b) is not reflected in (c). On the contrary, the cross-entropy in (d) reflects it, and treats false positives and false negatives differently. Best viewed in colors at high resolution.}
                                        \label{fig:gradientmap}
                                \end{figure}

                                In essence, the gradient map is a dynamically weighted negative of the ground truth $y$, as illustrated in Figure \ref{fig:gradientmap}.
                                This weight depends on the global intersection and union but is shared across all voxels of the same ground-truth class $k$.
                                This has two main consequences.

                                The first one is that as long as a single voxel is mis-classified (\emph{i.e.}, $\exists i: s_{\boldsymbol\theta}^{(i,k)} \neq y^{(i,k)}$), all the foreground voxels will have a non-negative gradients, even for voxels $j$ where $s_{\boldsymbol\theta}^{(j,k)} = y^{(j,k)} = 1$, as $\big(U^{(k)} - I^{(k)} = 0\big) \Longleftrightarrow \big(\Omega^{(k)}_y = \Omega^{(k)}_s\big)$.

                                The second consequence is that the gradient of the background pixels is $0$ only when the two segmentations have no overlap, $\big(I^{(k)} =0\big) \Longleftrightarrow \big(\Omega_y^{(k)} \cap \Omega_{s_{\boldsymbol\theta}}^{(k)} = \emptyset\big)$, while it will remain non-null for a perfect
                                segmentation
                                \footnote{We assume that $\Omega_y^{(k)}$ is non-empty. If this happens in practice the divide-by-zero can be avoided with a small $\epsilon$ that is added to the implementation (\emph{e.g.}, $\frac{-2 I^{(k)}}{U^{(k)}+\epsilon}$ ).}.

                                Those side-effects are likely not intentional from the design of the Dice coefficient and subsequent dice loss, but are counter-intuitive and surprising.
                                Ultimately, the dice loss encodes the ground-truth information in a rather crude way, losing a lot of nuance during the back-propagation process, while sometimes ``punishing'' good segmentation more harshly than bad ones.

                        \subsubsection{Upper bound}
                                We have $0 \leq 2 I^{(k)} \leq U^{(k)}$ (as $s_{\boldsymbol\theta}^{(i,k)},y^{(i,k)}\in [0, 1]$, so $y^{(i,k)}s_{\boldsymbol\theta}^{(i,k)} \leq s_{\boldsymbol\theta}^{(i,k)}$  and $y^{(i,k)}s_{\boldsymbol\theta}^{(i,k)} \leq y^{(i,k)}$) and therefore:
                                \begin{equation}
                                        \left|\frac{\partial \mathcal L_\text{DSC}}{\partial s_{\boldsymbol\theta}^{(i, k)}}\right| \leq \frac{2U^{(k)}}{\left(U^{(k)}\right)^2} = \frac{2}{U^{(k)}}.
                                \end{equation}
                                $U^{(k)}$ is bound to grow rather rapidly, as it is the sum of both the ground truth and network predictions over all voxels, and consequently, the magnitude of the dice loss gradient becomes quickly negligible. Going back to Figure \ref{fig:gradientmap}, the dynamic range (\emph{i.e.} the ratio between its maximum and minimum value) of $\mathcal L_\text{DSC}$ gradient is about 1.5dB, while it is 10dB for $\mathcal L_\text{CE}$.

                \subsection{Mimicking the dice loss}
                        One is left to ponder if the dynamic weight of \eqref{eq:dynamic_weight} is truly needed.
                        The core of the supervision seems to come not from the magnitude of the gradient, but its sign: a negative gradient pushes the probability \emph{up} (as this is a gradient descent), while a positive gradient pushes it \emph{down}.
                        As such, a similar effect can be achieved with a loss with a gradient of the form:
                        \begin{equation}
                                \label{eq:mimemap}
                                \frac{\partial \mathcal L_\text{Mime}}{\partial s_{\boldsymbol\theta}^{(i, k)}} := \begin{cases}
                                        - a & \text{if } y^{(i, k)} = 1 ,\\
                                        b & \text{otherwise,}
                                \end{cases}
                        \end{equation}
                        with $a,b > 0$. Their value can either be set empirically by the researcher, or be pre-computed based on (for instance) the dataset classes-distribution.

                        This would be further motivated by the relationship that exists between learning rate ($\eta$), loss weight ($\lambda$) (if any) and network gradient ($\nabla \boldsymbol\theta=\frac{\partial \mathcal L}{\partial s_{\boldsymbol\theta}}\frac{\partial s_{\boldsymbol\theta}}{\partial \boldsymbol\theta}$). Taking the update of a standard SGD: $\boldsymbol\theta_{t+1} = \boldsymbol\theta_t - \eta\lambda\nabla \boldsymbol\theta_t$,
                        a change in the magnitude of one element will have to be counter-balanced in the others to maintain similar network updates.

                        We can easily define a loss based on \eqref{eq:mimemap}:
                        \begin{equation}
                                \label{eq:l_rn}
                                \mathcal L_\text{Mime}(y, s_{\boldsymbol\theta}) := \boldsymbol \omega_y^\top\boldsymbol s_{\boldsymbol\theta} ,
                        \end{equation}
                        with $\boldsymbol \omega_y \in \mathbb R^{|\mathcal K||\Omega|}$ the flattened, pre-computed gradient map, and $\boldsymbol s_{\boldsymbol\theta} \in [0, 1]^{|\mathcal K||\Omega|}$ the flattened predicted probabilities.
                        Assuming that $\boldsymbol y \in \{0, 1\}^{|\mathcal K||\Omega|}$ is the flattened ground truth, we can trivially compute $\boldsymbol \omega_y = - \boldsymbol y a + (1 - \boldsymbol y) b$ with $a, b > 0$.

                \subsection{Further simplification for multi-class tasks}
                        A positive $b$ in \eqref{eq:mimemap} is useful to prevent false-positives in binary tasks, by pushing the predicted probabilities down outside of the object.
                        Not doing so would lead to trivial solutions, such as predicting the whole image as foreground.

                        However, those trivial solutions exist because there is a direct relationship between foreground and background in the binary case ($y^{(\cdot,0)} = 1 - y^{(\cdot, 1)}$, and vice versa).
                        This is not the case when $K \geq 2$, where classes have a more complex relationship: a trivial solution for one class is a really bad one for another one.
                        In a way, they all balance each others, creating an implicit regularization.
                        For those tasks, the loss can be simplified further, without a risk of catastrophic collapse:
                        \begin{equation}
                                \label{eq:l_rnt}
                                \mathcal L_\text{NM}(y, s_{\boldsymbol\theta}) := - \boldsymbol y^\top\boldsymbol s_{\boldsymbol\theta}.
                        \end{equation}

        \section{Experimental setup}
                \subsection{Compared losses}
                We perform experiments on several losses: cross-entropy ($\mathcal L_\text{CE}$, Eq. \eqref{eq:l_ce}), Dice ($\mathcal L_\text{DSC}$, Eq. \eqref{eq:l_dsc}) and two mimes ($\mathcal L_\text{Mime}$, Eq. \eqref{eq:l_rn}, and $\mathcal L_\text{NM}$, \eqref{eq:l_rnt}).
                For $\mathcal L_\text{Mime}$, while we tested different (usually successful) mimes, we stick here to the simplest version we tried, with $\boldsymbol\omega_y = - 2\boldsymbol y + 0.1$, for a final $a=-1.9$ and $b=0.1$. 

                Note that \emph{we are not} attempting here to find-out if one loss is better or the other. Rather, we want to test the hypothesis that the effectiveness of $\mathcal L_\text{DSC}$ comes from the signed gradient, rather its magnitude. We consider that the hypothesis would be validated if $\mathcal L_\text{Mime}$ and $\mathcal L_\text{NM}$ perform similarly, both in performances, convergence speed and visual segmentation.

                \subsection{Network architectures and optimizers}
                        To assess the generalizability of our findings, we experiment on two different network architectures (ENet \cite{paszke2016enet}, and a fully residual U-Net \cite{kervadec2021beyond}), trained with and without data augmentation, and experimented with two different optimizers: \textsc{Adam} \cite{kingma2014adam} (with $\beta=(0.99,0.999)$ and starting learning rate of $5\times10^{-4}$ ), and standard Stochastic Gradient Descent (SGD) (with a momentum of $0.9$, weight decay of $5\times10^{-4}$, and starting learning rate of $1\times10^{-2}$). We train for a fixed number of epochs, and the learning rate is halved if validation performances have not improved for 20 epochs.

                        Our code is implemented in PyTorch \cite{paszke2017automatic} and is anonymously available at \url{https://github.com/HKervadec/segmentation\_losses\#mime-loss}.

                \subsection{Datasets}
                        \paragraph{ACDC \cite{bernard2018deep}} is a dataset of cine-MRI of the heart, providing annotations at systole and diastole of the right-ventricle (RV), myocardium (\textsc{Myo}) and left-ventricle (LV) so that $K=3$.
                        The dataset contains 100 patients with different pathologies. We kept 10 patients for validation and 20 for testing.
                        For this dataset we use the ENet \cite{paszke2016enet} network, without data augmentation.

                        \paragraph{\textsc{Promise12} \cite{litjens2014evaluation}} is a challenge dataset of prostate MRI ($K=1$), coming from four centers with different scanners and acquisition protocols (resulting in significant variability in image contrast across the dataset).
                        In total, 50 cases with annotations are publicly available, of which we keep 5 for validation and 10 for testing.
                        We use the ResidualUNet with data augmentation for this task.

                \subsection{Metrics}
                        We report DSC (Eq. \eqref{eq:dsc}) on the testing set, and the validation DSC over time to assess convergence speed and stability. 
                        Moreover, we compare the testing calibration (as Classwise-Expected Calibration Error (ClECE) \cite{mukhoti2020calibrating}) of the different methods.

        \section{Results and discussion}
                \begin{table}[h]
                        \centering
                        \caption{Testing DSC (\%) with \emph{mean (standard deviation)} reported.}
                        \label{table:dsc}
                                \begin{tabular}{clccc|c|c}
                                        \toprule
                                        \multicolumn{2}{c}{\multirow{2}{*}{\diagbox{Loss}{Dataset}}}   & \multicolumn{4}{c|}{ACDC} & \textsc{Promise12} \\
                                        \cmidrule(lr){3-6}\cmidrule(lr){7-7}
                                        & & RV & \textsc{Myo} & LV & All & Prostate \\
                                        \cmidrule(lr){3-3}\cmidrule(lr){4-4}\cmidrule(lr){5-5}\cmidrule(lr){6-6}\cmidrule(lr){7-7}
                                       \multirow{4}{*}{\rotatebox{90}{\textsc{Adam}}} & $\mathcal L_\text{CE}$ & 81.7 (10.8) & 80.0 (07.2) & 89.4 (08.7) & 83.7 (09.9) & 87.1 (05.0) \\
                                        & $\mathcal L_\text{DSC}$ & 77.2 (13.6) & 79.2 (08.3) & 90.3 (06.3) & 82.2 (11.4) & 83.8 (07.2) \\
                                        & $\mathcal L_\text{Mime}$ & 82.6 (11.1) & 78.5 (08.4) & 87.8 (08.9) & 83.0 (10.3) & 83.4 (03.7) \\
                                        & $\mathcal L_\text{NM}$ & 81.5 (12.3) & 80.2 (07.3) & 90.9 (06.6) & 84.2 (10.3) & --- \\
                                        \hline
                                       \multirow{4}{*}{\rotatebox{90}{\textsc{SGD}}} & $\mathcal L_\text{CE}$ & 75.8 (14.6) & 76.6 (09.8) & 89.2 (08.1) & 80.6 (12.8) & 84.9 (08.3) \\
                                        & $\mathcal L_\text{DSC}$ & 76.3 (14.0) & 78.4 (08.3) & 90.2 (06.8) & 81.6 (11.9) & 84.8 (05.6) \\
                                        & $\mathcal L_\text{Mime}$ & 78.3 (13.7) & 77.8 (084) & 88.6 (06.9) & 81.6 (11.3) & 84.0 (03.9) \\
                                        & $\mathcal L_\text{NM}$ & 79.2 (14.3) & 78.3 (08.9) & 88.7 (07.4) & 82.1 (11.6) & --- \\
                                        \bottomrule
                                \end{tabular}
                \end{table}

                \pgfplotsset{
                    compat=newest,
                    /pgfplots/legend image code/.code={%
                        \draw[ultra thick, #1]
                            plot coordinates {
                                (0cm,0cm)
                                (0.3cm,.1cm)
                                (0.6cm,0cm)
                                (0.9cm,-.1cm)
                                (1.2cm,0cm)%
                            };
                    },
                }
                \begin{figure}[h]
                        \centering
                        \pgfplotslegendfromname{commonadam}
                        \begin{tikzpicture}
                                \begin{axis}[
                                        name=acdcdsc,
                                        width=0.49\linewidth,
                                        height=0.35\linewidth,
                                        at={(0,0)},
                                        ymin=0,
                                        ymax=1,
                                        xmin=0,
                                        xmax=199,
                                        ymajorgrids=true,
                                        ylabel={DSC (\textsc{Adam} opt.)},
                                        every axis y label/.style={
                                                at={(ticklabel cs:0.5)},rotate=90,anchor=near ticklabel,
                                        },
                                        every axis x label/.style={
                                                at={(ticklabel cs:0.5)},anchor=near ticklabel,
                                        },
                                        xtick={0,100,199},
                                        xticklabels={0,100,200},
                                        ytick={0,0.1,...,1},
                                        yticklabels={0.0,0.1,0.2,0.3,0.4,0.5,0.6,0.7,0.8,0.9,1.0},
                                        legend pos=outer north east,
                                        legend cell align=left,
                                        tick label style={font=\scriptsize},
                                        label style={font=\scriptsize},
                                        title style={font=\scriptsize},
                                        legend to name=commonadam,
                                        legend columns=4,
                                ]
                                        \addplot[color=MidnightBlue, semithick] table [x=epoch, y=ce-mean, col sep=comma, mark=none] {constrained_cnn-221209-592be6a-huehuecoyotl-acdc_ipmi/val_3d_dsc.csv};
                                        \addlegendentry{$\mathcal L_\text{CE}$}

                                        \addplot[color=Cerulean, semithick] table [x=epoch, y=dice-mean, col sep=comma, mark=none] {constrained_cnn-221209-592be6a-huehuecoyotl-acdc_ipmi/val_3d_dsc.csv};
                                        \addlegendentry{$\mathcal L_\text{DSC}$}

                                        \addplot[color=Green, semithick] table [x=epoch, y=rn_tricktransform-mean, col sep=comma, mark=none] {constrained_cnn-221209-592be6a-huehuecoyotl-acdc_ipmi/val_3d_dsc.csv};
                                        \addlegendentry{$\mathcal L_\text{Mime}$}

                                        \addplot[color=BrickRed, semithick] table [x=epoch, y=rn-mean, col sep=comma, mark=none] {constrained_cnn-221209-592be6a-huehuecoyotl-acdc_ipmi/val_3d_dsc.csv};
                                        \addlegendentry{$\mathcal L_\text{NM}$}
                                \end{axis}
                                \begin{axis}[
                                        name=promisedsc,
                                        width=0.49\linewidth,
                                        height=0.35\linewidth,
                                        at={($(acdcdsc.east)+(2em,0)$)},
                                        anchor=west,
                                        ymin=0,
                                        ymax=1,
                                        xmin=0,
                                        xmax=299,
                                        ymajorgrids=true,
                                        every axis y label/.style={
                                                at={(ticklabel cs:0.5)},rotate=90,anchor=near ticklabel,
                                        },
                                        every axis x label/.style={
                                                at={(ticklabel cs:0.5)},anchor=near ticklabel,
                                        },
                                        xtick={0,100,200,299},
                                        xticklabels={0,100,200,300},
                                        ytick={0,0.1,...,1},
                                        yticklabels={,,,,,,,,,,},
                                        legend cell align=left,
                                        tick label style={font=\scriptsize},
                                        label style={font=\scriptsize},
                                        title style={font=\scriptsize},
                                ]
                                        \addplot[color=MidnightBlue, semithick] table [x=epoch, y=ce, col sep=comma, mark=none] {constrained_cnn-221210-592be6a-huehuecoyotl-prostate_ipmi/val_3d_dsc.csv};

                                        \addplot[color=Cerulean, semithick] table [x=epoch, y=dice, col sep=comma, mark=none] {constrained_cnn-221210-592be6a-huehuecoyotl-prostate_ipmi/val_3d_dsc.csv};

                                        \addplot[color=Green, semithick] table [x=epoch, y=rn_tricktransform, col sep=comma, mark=none] {constrained_cnn-221210-592be6a-huehuecoyotl-prostate_ipmi/val_3d_dsc.csv};
                                \end{axis}
                        \end{tikzpicture}
                        \begin{tikzpicture}
                                \begin{axis}[
                                        name=acdcdscsgd,
                                        width=0.49\linewidth,
                                        height=0.35\linewidth,
                                        at={(0,0)},
                                        anchor=north,
                                        ymin=0,
                                        ymax=1,
                                        xmin=0,
                                        xmax=199,
                                        ymajorgrids=true,
                                        ylabel={DSC (SGD opt.)},
                                        every axis y label/.style={
                                                at={(ticklabel cs:0.5)},rotate=90,anchor=near ticklabel,
                                        },
                                        every axis x label/.style={
                                                at={(ticklabel cs:0.5)},anchor=near ticklabel,
                                        },
                                        xlabel=Epoch (ACDC),
                                        xtick={0,100,199},
                                        xticklabels={0,100,200},
                                        ytick={0,0.1,...,1},
                                        yticklabels={0.0,0.1,0.2,0.3,0.4,0.5,0.6,0.7,0.8,0.9,1.0},
                                        legend pos=outer north east,
                                        legend cell align=left,
                                        tick label style={font=\scriptsize},
                                        label style={font=\scriptsize},
                                        title style={font=\scriptsize},
                                ]
                                        \addplot[color=MidnightBlue, semithick] table [x=epoch, y=ce_sgd-mean, col sep=comma, mark=none] {constrained_cnn-221209-592be6a-huehuecoyotl-acdc_ipmi/val_3d_dsc.csv};

                                        \addplot[color=Cerulean, semithick] table [x=epoch, y=dice_sgd-mean, col sep=comma, mark=none] {constrained_cnn-221209-592be6a-huehuecoyotl-acdc_ipmi/val_3d_dsc.csv};

                                        \addplot[color=Green, semithick] table [x=epoch, y=rn_tricktransform_sgd10-mean, col sep=comma, mark=none] {constrained_cnn-221209-592be6a-huehuecoyotl-acdc_ipmi/val_3d_dsc.csv};

                                        \addplot[color=BrickRed, semithick] table [x=epoch, y=rn_sgd-mean, col sep=comma, mark=none] {constrained_cnn-221209-592be6a-huehuecoyotl-acdc_ipmi/val_3d_dsc.csv};
                                \end{axis}
                                \begin{axis}[
                                        name=promisedscsgd,
                                        width=0.49\linewidth,
                                        height=0.35\linewidth,
                                        at={($(acdcdscsgd.east)+(2em,0)$)},
                                        anchor=west,
                                        ymin=0,
                                        ymax=1,
                                        xmin=0,
                                        xmax=299,
                                        ymajorgrids=true,
                                        every axis y label/.style={
                                                at={(ticklabel cs:0.5)},rotate=90,anchor=near ticklabel,
                                        },
                                        every axis x label/.style={
                                                at={(ticklabel cs:0.5)},anchor=near ticklabel,
                                        },
                                        xlabel=Epoch (\textsc{Promise12}),
                                        xtick={0,100,200,299},
                                        xticklabels={0,100,200,300},
                                        ytick={0,0.1,...,1},
                                        yticklabels={,,,,,,,,,,},
                                        legend pos=outer north east,
                                        legend cell align=left,
                                        tick label style={font=\scriptsize},
                                        label style={font=\scriptsize},
                                        title style={font=\scriptsize},
                                ]
                                        \addplot[color=MidnightBlue, semithick] table [x=epoch, y=ce_sgd, col sep=comma, mark=none] {constrained_cnn-221210-592be6a-huehuecoyotl-prostate_ipmi/val_3d_dsc.csv};

                                        \addplot[color=Cerulean, semithick] table [x=epoch, y=dice_sgd, col sep=comma, mark=none] {constrained_cnn-221210-592be6a-huehuecoyotl-prostate_ipmi/val_3d_dsc.csv};

                                        \addplot[color=Green, semithick] table [x=epoch, y=rn_tricktransform_sgd10, col sep=comma, mark=none] {constrained_cnn-221210-592be6a-huehuecoyotl-prostate_ipmi/val_3d_dsc.csv};
                                \end{axis}
                        \end{tikzpicture}
                        \caption{Validation DSC over time for both datasets with the \textsc{Adam} optimizer.}
                        \label{fig:validation_metrics_plot_adam}
                \end{figure}
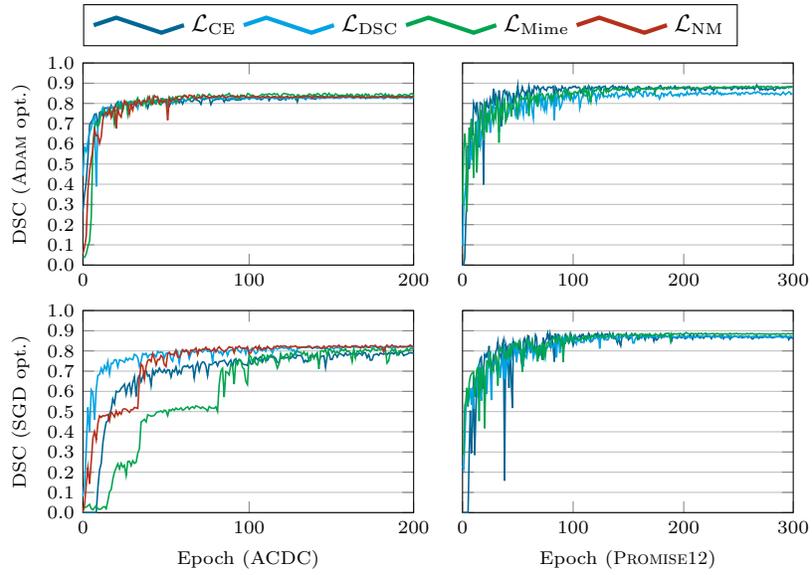

                \begin{figure}[h]
                        \centering
                        \begin{subfigure}[t]{0.2\textwidth}
                                \centering
                                \includegraphics[width=\textwidth]{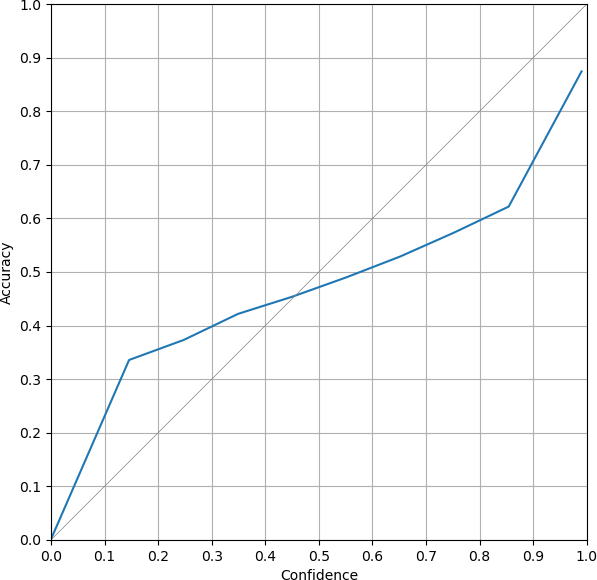}
                                \caption{$\mathcal L_\text{CE}$}
                        \end{subfigure}
                        \begin{subfigure}[t]{0.2\textwidth}
                                \centering
                                \includegraphics[width=\textwidth]{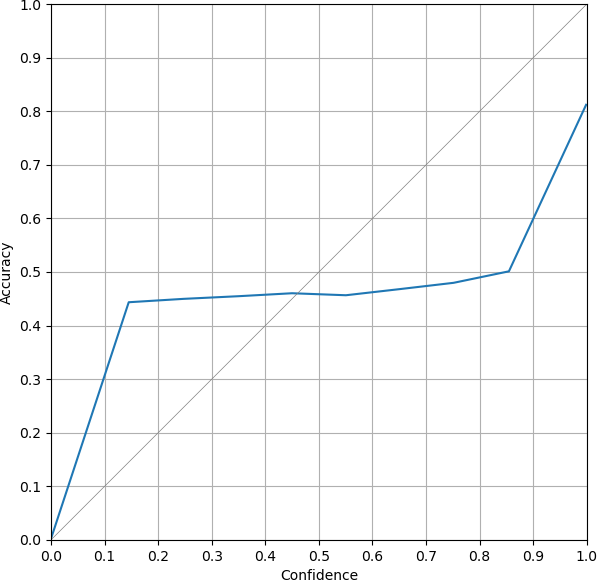}
                                \caption{$\mathcal L_\text{DSC}$}
                        \end{subfigure}
                        \begin{subfigure}[t]{0.2\textwidth}
                                \centering
                                \includegraphics[width=\textwidth]{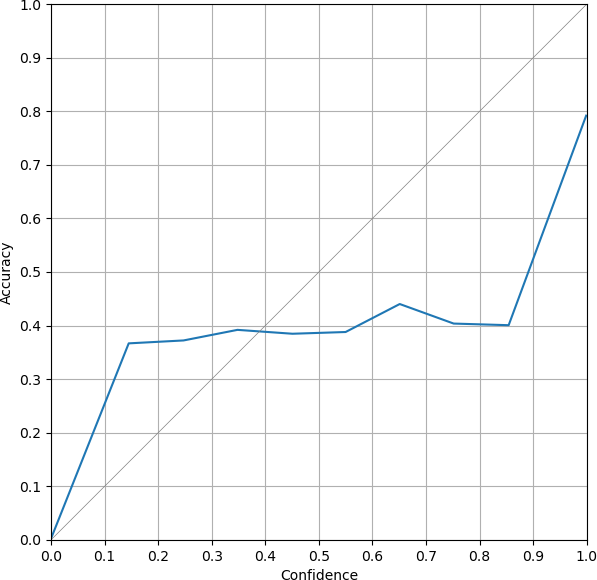}
                                \caption{$\mathcal L_\text{Mime}$}
                        \end{subfigure}
                        \begin{subfigure}[t]{0.2\textwidth}
                                \centering
                                \includegraphics[width=\textwidth]{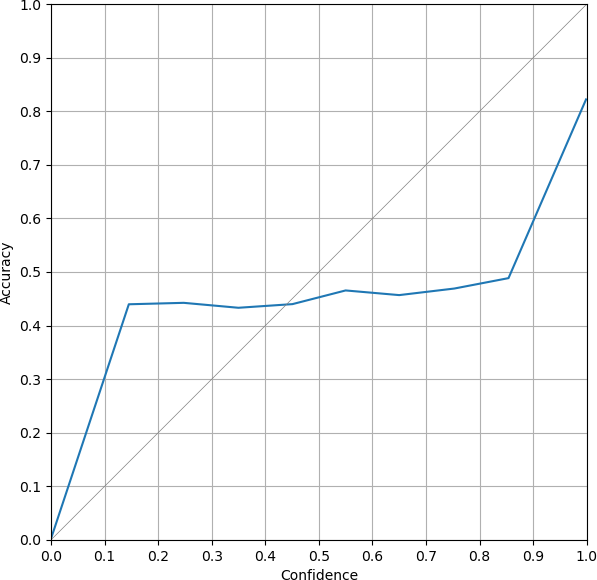}
                                \caption{$\mathcal L_\text{NM}$}
                        \end{subfigure}
                        \caption{ClECE (Class-wise Expected Calibration Error) for the LV class of ACDC.}
                        \label{fig:calibration}
                \end{figure}



                \begin{figure}[h]
                        \centering
                        \begin{subfigure}[t]{0.19\linewidth}
                                \centering
                                \includegraphics[width=\textwidth, trim=30 30 30 30, clip=true]{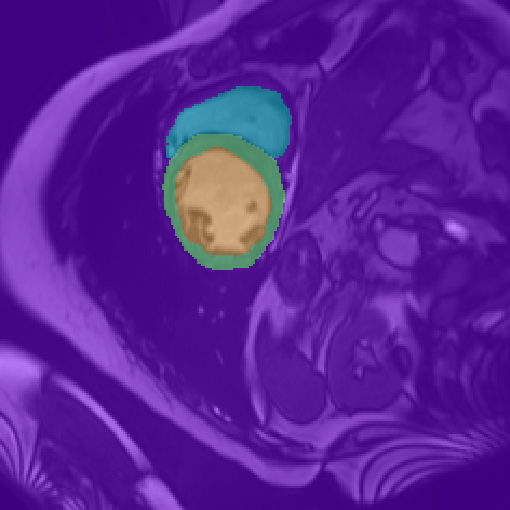}
                        \end{subfigure}
                        \begin{subfigure}[t]{0.19\linewidth}
                                \centering
                                \includegraphics[width=\textwidth, trim=30 30 30 30, clip=true]{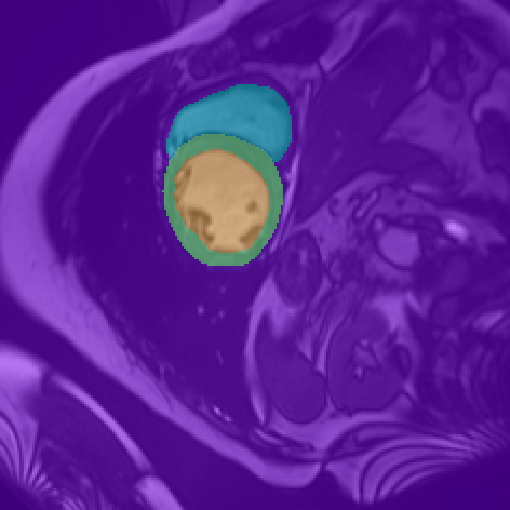}
                        \end{subfigure}
                        \begin{subfigure}[t]{0.19\linewidth}
                                \centering
                                \includegraphics[width=\textwidth, trim=30 30 30 30, clip=true]{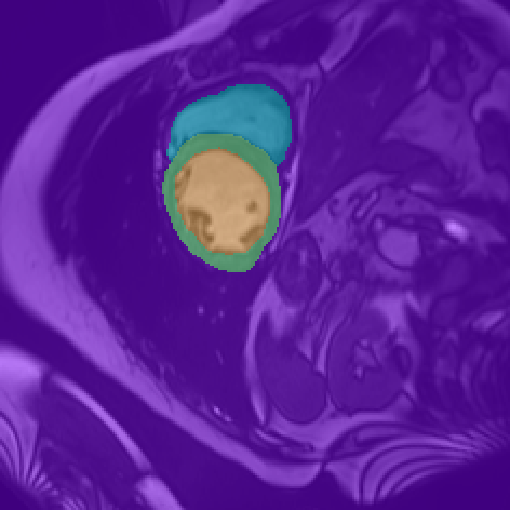}
                        \end{subfigure}
                        \begin{subfigure}[t]{0.19\linewidth}
                                \centering
                                \includegraphics[width=\textwidth, trim=30 30 30 30, clip=true]{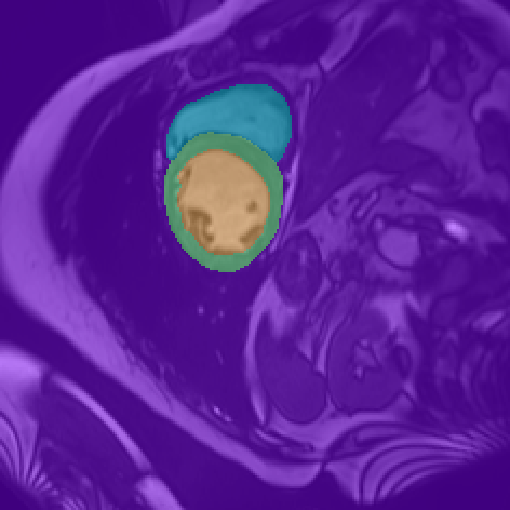}
                        \end{subfigure}
                        \begin{subfigure}[t]{0.19\linewidth}
                                \centering
                                \includegraphics[width=\textwidth, trim=30 30 30 30, clip=true]{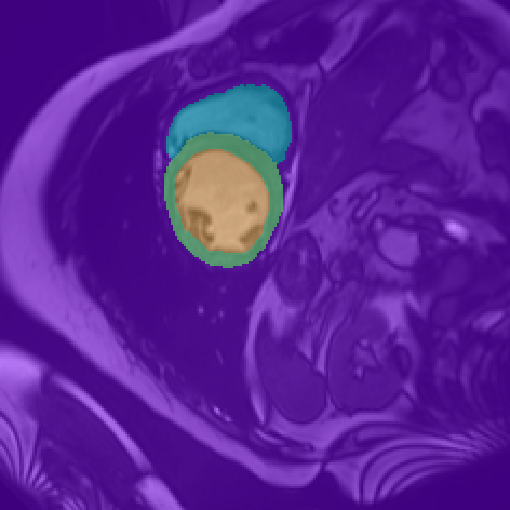}
                        \end{subfigure}
                        \\
                        \begin{subfigure}[t]{0.19\linewidth}
                                \centering
                                \includegraphics[width=\textwidth, trim=30 30 30 30, clip=true]{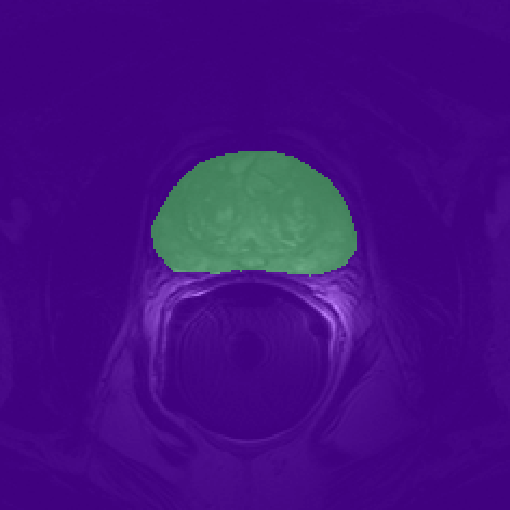}
                                \caption{$y$}
                        \end{subfigure}
                        \begin{subfigure}[t]{0.19\linewidth}
                                \centering
                                \includegraphics[width=\textwidth, trim=30 30 30 30, clip=true]{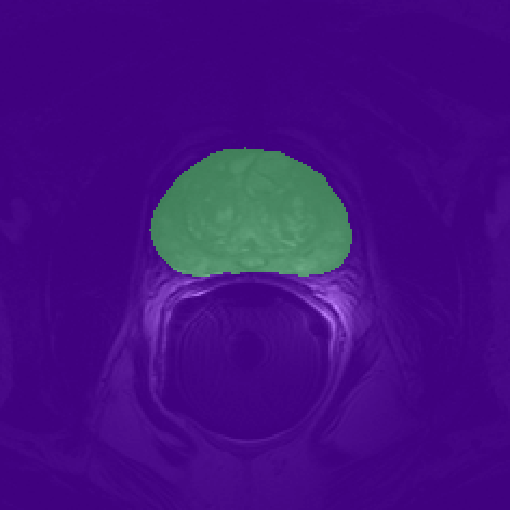}
                                \caption{$\mathcal L_\text{CE}$}
                        \end{subfigure}
                        \begin{subfigure}[t]{0.19\linewidth}
                                \centering
                                \includegraphics[width=\textwidth, trim=30 30 30 30, clip=true]{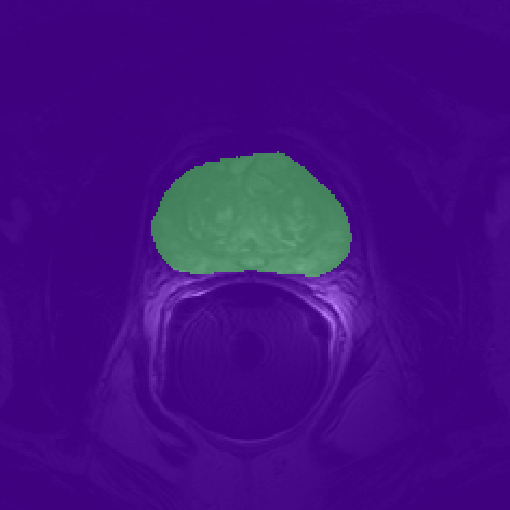}
                                \caption{$\mathcal L_\text{DSC}$}
                        \end{subfigure}
                        \begin{subfigure}[t]{0.19\linewidth}
                                \centering
                                \includegraphics[width=\textwidth, trim=30 30 30 30, clip=true]{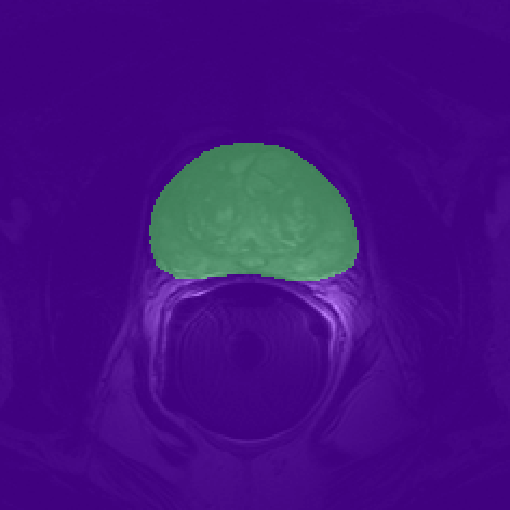}
                                \caption{$\mathcal L_\text{Mime}$}
                        \end{subfigure}
                        \begin{subfigure}[t]{0.19\linewidth}
                                \centering
                                Not applicable
                                \caption{$\mathcal L_\text{NM}$}
                        \end{subfigure}
                        \caption{Visual comparison on the testing sets when using \textsc{Adam}. In the manuscript we cropped the image around the object for visibility.}
                        \label{fig:visual_main}
                \end{figure}

                We report testing DSC for both datasets in Table \ref{table:dsc}. As we can see, all losses perform similarly. 
                Figure \ref{fig:validation_metrics_plot_adam} shows that all methods converge at a similar speed when using the \textsc{Adam} optimizer.
                When using SGD, some losses can get stuck for a while, before being likely ``un-blocked'' by the halving of the learning rate.
                In Figure \ref{fig:calibration}, we can see that both the $\mathcal L_\text{DSC}$ and its mimes have a similar calibration, with a flat accuracy for the confidence bins ranging from $0.1$ to $0.8$. (We report only one class due to space constraints, but this finding carried across other settings.)
                Qualitative results are visible in Figure \ref{fig:visual_main}, with extra examples in the supplementary material \ref{appendix:sec:extraqualitative}.
                While there is some variance between methods (due to the inherent variability of stochastic optimization), they all segment the objects satisfactorily.

                Another notable (but hardly new) indirect finding, is that \textsc{Adam} as an optimizer is much easier to work with.
                SGD required much more tuning of its learning weight and other parameters to work (affecting all losses, not only our mimes), with hyper-parameters not carrying across settings.
                This is consistent with our hypothesis, that the supervision of the DSC comes not from the gradient magnitude, but its sign.
                With \textsc{Adam}, and its partly automatic rescaling of the gradients, the manual tuning of learning rate and loss weight is not required. 

        \section{Conclusion and future works}
                We have shown, both analytically and empirically, that it is possible to mimic the dice loss with a simple negative of the ground truth.
                This indicates that supervision comes mostly from the gradient sign, and much less than its magnitude.
                This interesting finding should help to interpret and diagnose results when tuning losses for segmentation networks, and guide researchers when evaluating or designing novel segmentation losses.

                Those initial findings also point to many future and exciting new works.
                For instance, the simpler mime might be more straightforward to adapt for imbalanced tasks \cite{sudre2017generalised}. Interestingly (and ironically, in a way), the results of this paper point to a previously unseen relationship between Dice loss and Boundary loss \cite{kervadec2021boundary}.
                Another area to investigate is the behavior with respect to sub-patching. While \emph{computable}, the definition of Dice loss makes less semantic sense when applied to only a sub-patch of an image---while potentially causing divide by 0 errors on empty patches---, but this is not the case with a mime loss.

        \clearpage
        \bibliographystyle{splncs04}
        \bibliography{main}

\begin{thebibliography}{10}
\providecommand{\url}[1]{\texttt{#1}}
\providecommand{\urlprefix}{URL }
\providecommand{\doi}[1]{https://doi.org/#1}

\bibitem{bernard2018deep}
Bernard, O., Lalande, A., Zotti, C., Cervenansky, F., Yang, X., Heng, P.A.,
  Cetin, I., Lekadir, K., Camara, O., Ballester, M.A.G., et~al.: Deep learning
  techniques for automatic mri cardiac multi-structures segmentation and
  diagnosis: is the problem solved? IEEE transactions on medical imaging
  \textbf{37}(11),  2514--2525 (2018)

\bibitem{eelbode2020optimization}
Eelbode, T., Bertels, J., Berman, M., Vandermeulen, D., Maes, F., Bisschops,
  R., Blaschko, M.B.: Optimization for medical image segmentation: theory and
  practice when evaluating with dice score or jaccard index. IEEE Transactions
  on Medical Imaging  \textbf{39}(11),  3679--3690 (2020)

\bibitem{isensee2021nnu}
Isensee, F., Jaeger, P.F., Kohl, S.A., Petersen, J., Maier-Hein, K.H.: nnu-net:
  a self-configuring method for deep learning-based biomedical image
  segmentation. Nature methods  \textbf{18}(2),  203--211 (2021)

\bibitem{kervadec2021beyond}
Kervadec, H., Bahig, H., Letourneau-Guillon, L., Dolz, J., Ayed, I.B.: Beyond
  pixel-wise supervision: semantic segmentation with higher-order shape
  descriptors. In: Medical Imaging with Deep Learning. vol.~133, pp. 354--368.
  PMLR (2021)

\bibitem{kervadec2021boundary}
Kervadec, H., Bouchtiba, J., Desrosiers, C., Granger, E., Dolz, J., Ayed, I.B.:
  Boundary loss for highly unbalanced segmentation. Medical Image Analysis
  \textbf{67},  101851 (2021)

\bibitem{kingma2014adam}
Kingma, D.P., Ba, J.: Adam: A method for stochastic optimization. arXiv
  preprint arXiv:1412.6980  (2014)

\bibitem{litjens2014evaluation}
Litjens, G., Toth, R., van~de Ven, W., Hoeks, C., Kerkstra, S., van Ginneken,
  B., Vincent, G., Guillard, G., Birbeck, N., Zhang, J., et~al.: Evaluation of
  prostate segmentation algorithms for mri: the promise12 challenge. Medical
  image analysis  \textbf{18}(2),  359--373 (2014)

\bibitem{ma2021loss}
Ma, J., Chen, J., Ng, M., Huang, R., Li, Y., Li, C., Yang, X., Martel, A.L.:
  Loss odyssey in medical image segmentation. Medical Image Analysis
  \textbf{71},  102035 (2021)

\bibitem{maier2022metrics}
Maier-Hein, L., Menze, B., et~al.: Metrics reloaded: Pitfalls and
  recommendations for image analysis validation. arXiv. org (2206.01653) (2022)

\bibitem{milletari2016v}
Milletari, F., Navab, N., Ahmadi, S.A.: V-net: Fully convolutional neural
  networks for volumetric medical image segmentation. In: 2016 fourth
  international conference on 3D vision (3DV). pp. 565--571. IEEE (2016)

\bibitem{mukhoti2020calibrating}
Mukhoti, J., Kulharia, V., Sanyal, A., Golodetz, S., Torr, P., Dokania, P.:
  Calibrating deep neural networks using focal loss. Advances in Neural
  Information Processing Systems  \textbf{33},  15288--15299 (2020)

\bibitem{nordstrom2020calibrated}
Nordstr{\"o}m, M., Bao, H., L{\"o}fman, F., Hult, H., Maki, A., Sugiyama, M.:
  Calibrated surrogate maximization of dice. In: Medical Image Computing and
  Computer Assisted Intervention--MICCAI 2020: 23rd International Conference,
  Lima, Peru, October 4--8, 2020, Proceedings, Part IV 23. pp. 269--278.
  Springer (2020)

\bibitem{paszke2016enet}
Paszke, A., Chaurasia, A., Kim, S., Culurciello, E.: Enet: A deep neural
  network architecture for real-time semantic segmentation. arXiv preprint
  arXiv:1606.02147  (2016)

\bibitem{paszke2017automatic}
Paszke, A., Gross, S., Chintala, S., Chanan, G., Yang, E., DeVito, Z., Lin, Z.,
  Desmaison, A., Antiga, L., Lerer, A.: Automatic differentiation in pytorch
  (2017)

\bibitem{sudre2017generalised}
Sudre, C.H., Li, W., Vercauteren, T., Ourselin, S., Jorge~Cardoso, M.:
  Generalised dice overlap as a deep learning loss function for highly
  unbalanced segmentations. In: Deep learning in medical image analysis and
  multimodal learning for clinical decision support, pp. 240--248. Springer
  (2017)

\end{thebibliography}

        \clearpage
        \appendix
        \setcounter{page}{1}
        \section{Notation summary}
                \label{sec:notation}

                \begin{align*}
                        \Omega &\subset \mathbb R^D & D\text{-dimensional image space} \\
                        \mathcal K &= \{0: \texttt{background},  ..., K: \texttt{classK} \} & \text{set of classes to segment;} \\
                        &K & \text{number of object classes;} \\
                        y^{(\centerdot,\centerdot)}&: (\Omega \times \mathcal K) \rightarrow \{0, 1\} & \text{one-hot encoded label;} \\
                        s^{(\centerdot,\centerdot)}&: (\Omega \times \mathcal K) \rightarrow \{0, 1\} & \text{predicted segmentation;} \\
                        &\boldsymbol\theta & \text{network parameters;} \\
                        s_{\boldsymbol\theta}^{(\centerdot,\centerdot)}&: (\Omega \times \mathcal K) \rightarrow [0, 1] & \text{predicted probabilities;} \\
                        \Omega_y^{(k)} &= \{i \in \Omega | y^{(i, k)}=1\} \subseteq \Omega & \text{subset of $\Omega$ of class $k$;} \\
                        \Omega_s^{(k)} &= \{i \in \Omega | s^{(i, k)}=1\} \subseteq \Omega & \text{subset of $\Omega$ predicted as $k$;} \\
                        I^{(k)} &= \sum_{i \in \Omega}y^{(i,k)}s_{\boldsymbol\theta}^{(i,k)} & \text{intersection;} \\
                        U^{(k)} &= \sum_{i \in \Omega}\left[y^{(i,k)} + s_{\boldsymbol\theta}^{(i,k)}\right] & \text{union;} \\
                        a ,b &\in \mathbb R & \text{mime loss weights;}\\
                        \exists i&: \texttt A & \text{it exists $i$ such as \texttt{A} is true;} \\
                        \texttt{A} &\Longleftrightarrow \texttt{B} & \text{\texttt{A} true if and only if \texttt{B};} \\
                        &\emptyset & \text{empty set;} \\
                        \nabla \boldsymbol\theta&=\frac{\partial \mathcal L}{\partial s_{\boldsymbol\theta}}\frac{\partial s_{\boldsymbol\theta}}{\partial \boldsymbol\theta} & \text{network gradient;} \\
                        &\eta & \text{learning rate;} \\
                        &\lambda & \text{optional loss weight;} \\
                        &\boldsymbol\theta_t & \text{parameters at iteration $t$;} \\
                        \boldsymbol \omega_y & \in \mathbb R^{|\mathcal K||\Omega|} & \text{flattened weight map matrix;} \\
                        \boldsymbol y & \in \mathbb \{0, 1\}^{|\mathcal K||\Omega|} & \text{flattened ground-truth matrix;} \\
                        \boldsymbol s_{\boldsymbol\theta} & \in [0, 1]^{|\mathcal K||\Omega|} & \text{flattened probabilities matrix;} \\
                        &\beta & \text{\textsc{Adam} main hyper-parameter.}
                \end{align*}

        \clearpage
        \section{Extra qualitative examples}
                \label{appendix:sec:extraqualitative}

                \begin{figure}[h]
                        \centering
                        \begin{subfigure}[t]{0.19\linewidth}
                                \centering
                                \includegraphics[width=\textwidth, trim=30 30 30 30, clip=true]{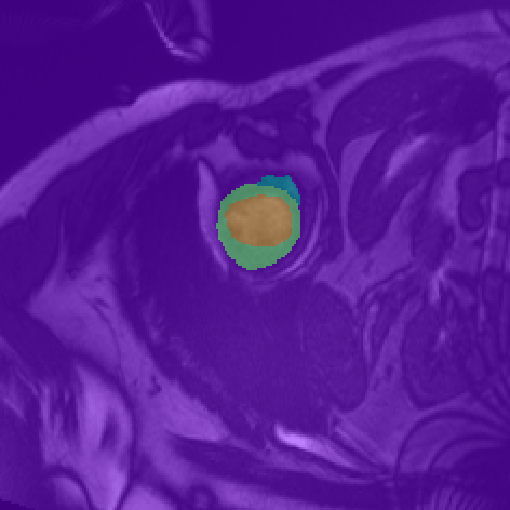}
                        \end{subfigure}
                        \begin{subfigure}[t]{0.19\linewidth}
                                \centering
                                \includegraphics[width=\textwidth, trim=30 30 30 30, clip=true]{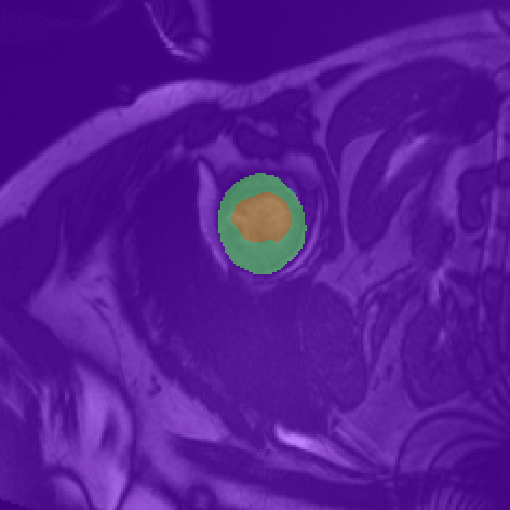}
                        \end{subfigure}
                        \begin{subfigure}[t]{0.19\linewidth}
                                \centering
                                \includegraphics[width=\textwidth, trim=30 30 30 30, clip=true]{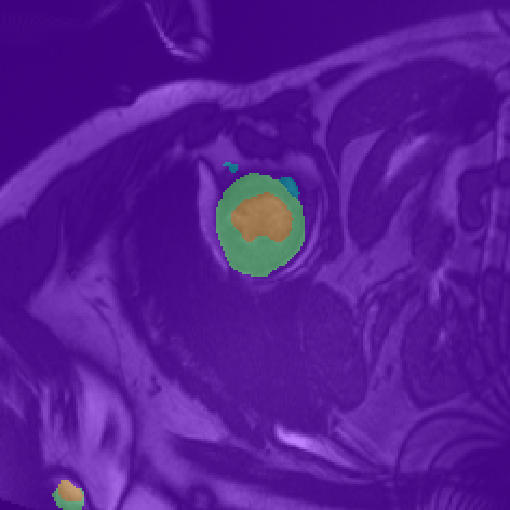}
                        \end{subfigure}
                        \begin{subfigure}[t]{0.19\linewidth}
                                \centering
                                \includegraphics[width=\textwidth, trim=30 30 30 30, clip=true]{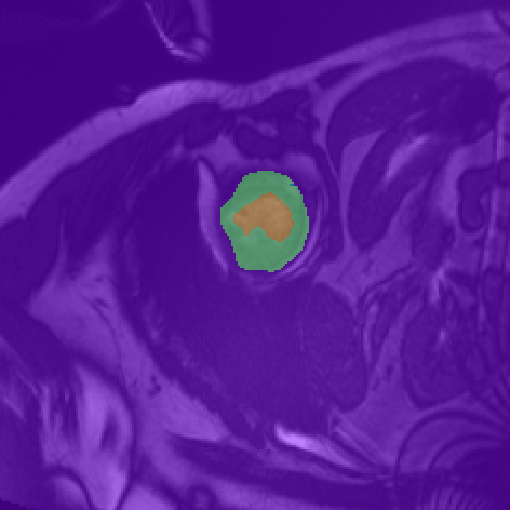}
                        \end{subfigure}
                        \begin{subfigure}[t]{0.19\linewidth}
                                \centering
                                \includegraphics[width=\textwidth, trim=30 30 30 30, clip=true]{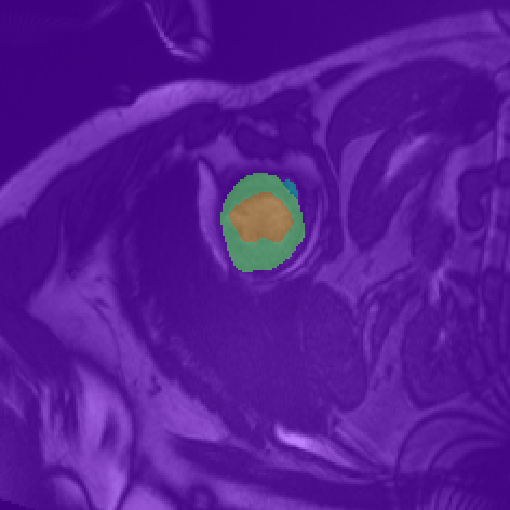}
                        \end{subfigure}
                        \\
                        \begin{subfigure}[t]{0.19\linewidth}
                                \centering
                                \includegraphics[width=\textwidth, trim=30 30 30 30, clip=true]{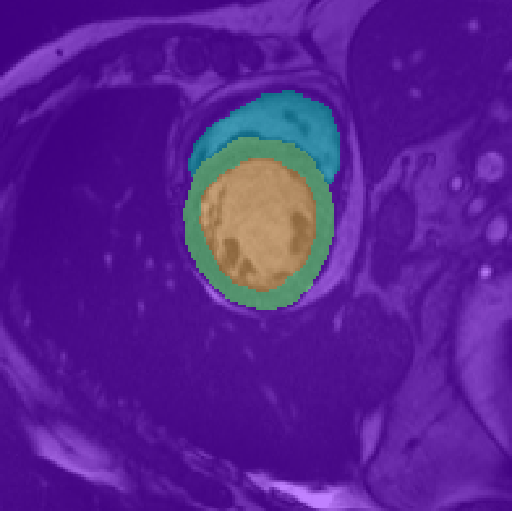}
                        \end{subfigure}
                        \begin{subfigure}[t]{0.19\linewidth}
                                \centering
                                \includegraphics[width=\textwidth, trim=30 30 30 30, clip=true]{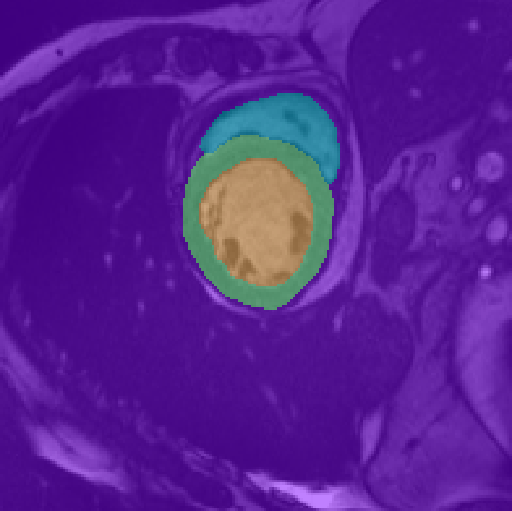}
                        \end{subfigure}
                        \begin{subfigure}[t]{0.19\linewidth}
                                \centering
                                \includegraphics[width=\textwidth, trim=30 30 30 30, clip=true]{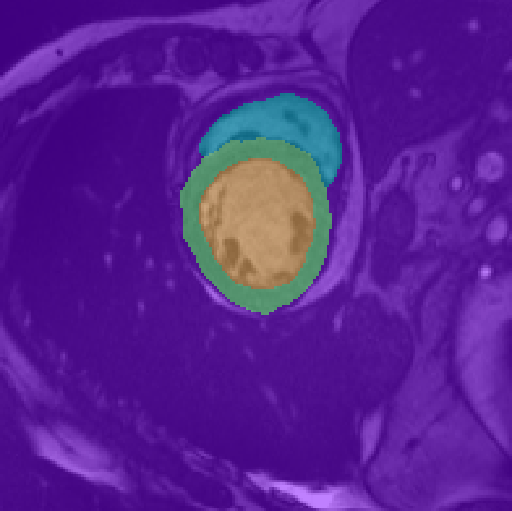}
                        \end{subfigure}
                        \begin{subfigure}[t]{0.19\linewidth}
                                \centering
                                \includegraphics[width=\textwidth, trim=30 30 30 30, clip=true]{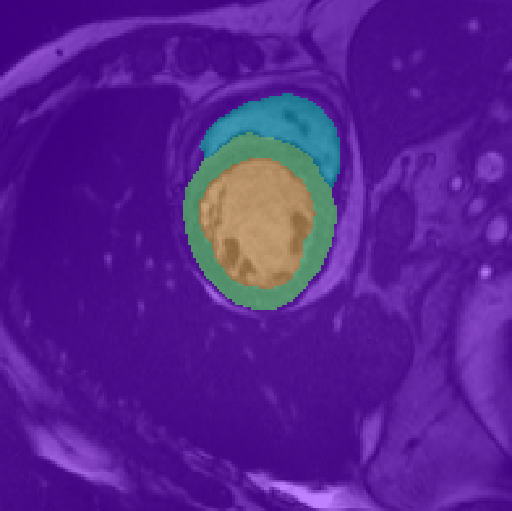}
                        \end{subfigure}
                        \begin{subfigure}[t]{0.19\linewidth}
                                \centering
                                \includegraphics[width=\textwidth, trim=30 30 30 30, clip=true]{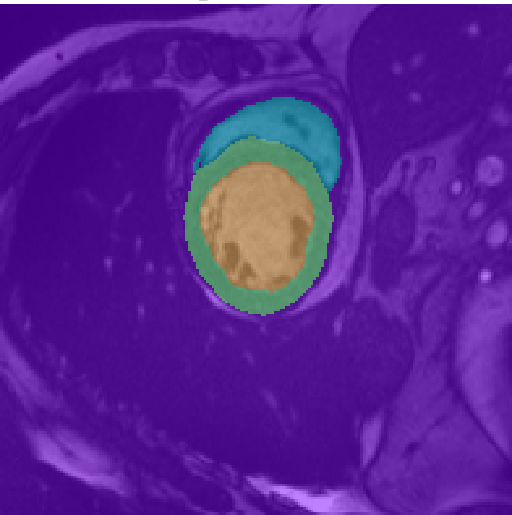}
                        \end{subfigure}
                        \\
                        \begin{subfigure}[t]{0.19\linewidth}
                                \centering
                                \includegraphics[width=\textwidth, trim=30 30 30 30, clip=true]{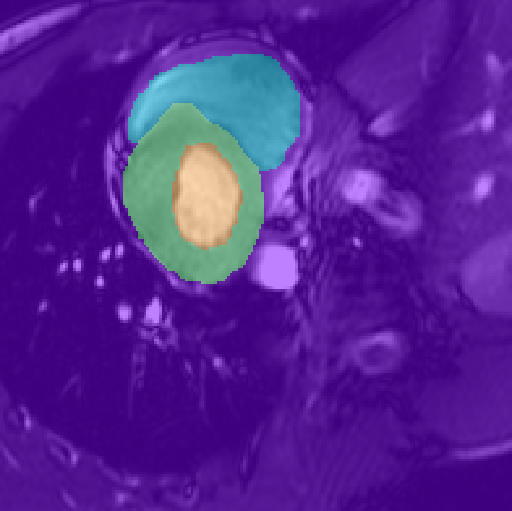}
                        \end{subfigure}
                        \begin{subfigure}[t]{0.19\linewidth}
                                \centering
                                \includegraphics[width=\textwidth, trim=30 30 30 30, clip=true]{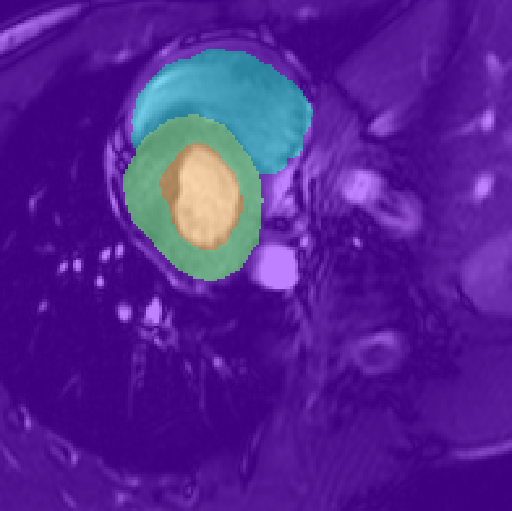}
                        \end{subfigure}
                        \begin{subfigure}[t]{0.19\linewidth}
                                \centering
                                \includegraphics[width=\textwidth, trim=30 30 30 30, clip=true]{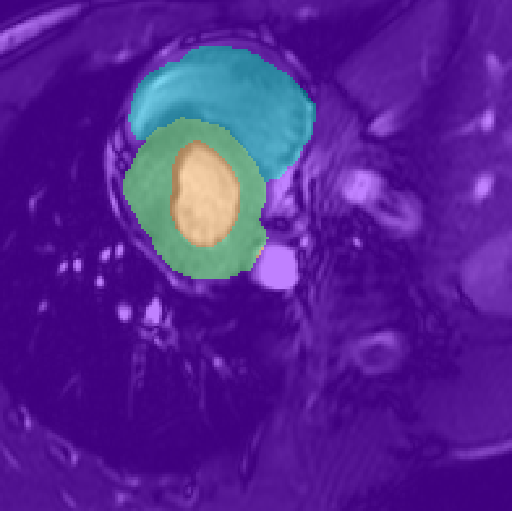}
                        \end{subfigure}
                        \begin{subfigure}[t]{0.19\linewidth}
                                \centering
                                \includegraphics[width=\textwidth, trim=30 30 30 30, clip=true]{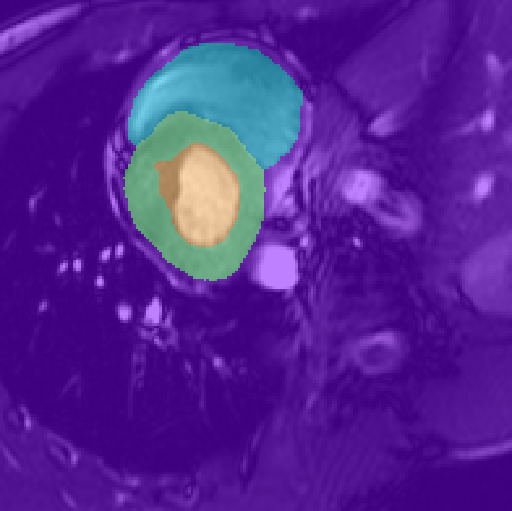}
                        \end{subfigure}
                        \begin{subfigure}[t]{0.19\linewidth}
                                \centering
                                \includegraphics[width=\textwidth, trim=30 30 30 30, clip=true]{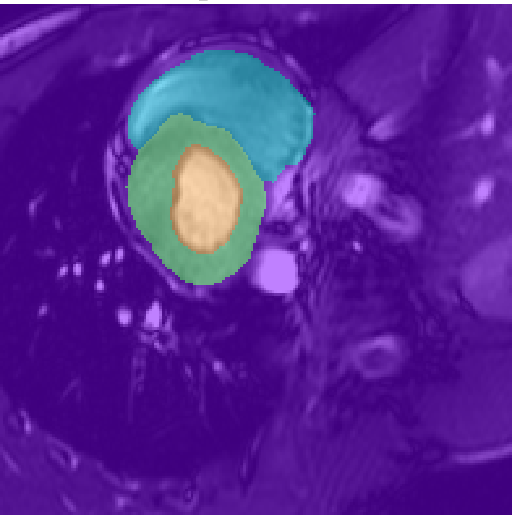}
                        \end{subfigure}
                        \\
                        \begin{subfigure}[t]{0.19\linewidth}
                                \centering
                                \includegraphics[width=\textwidth, trim=30 30 30 30, clip=true]{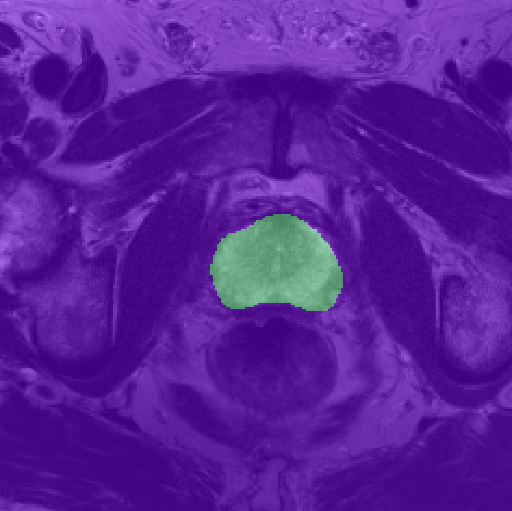}
                        \end{subfigure}
                        \begin{subfigure}[t]{0.19\linewidth}
                                \centering
                                \includegraphics[width=\textwidth, trim=30 30 30 30, clip=true]{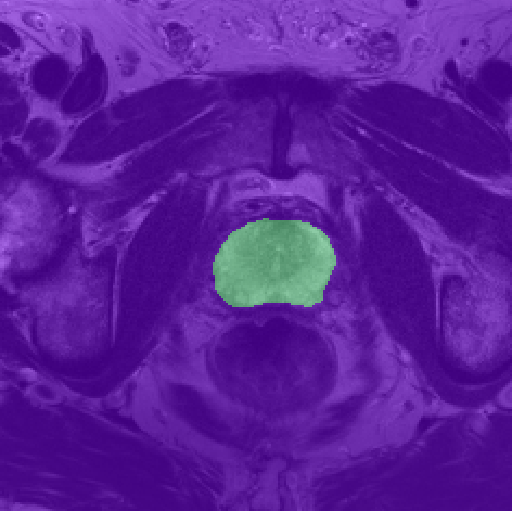}
                        \end{subfigure}
                        \begin{subfigure}[t]{0.19\linewidth}
                                \centering
                                \includegraphics[width=\textwidth, trim=30 30 30 30, clip=true]{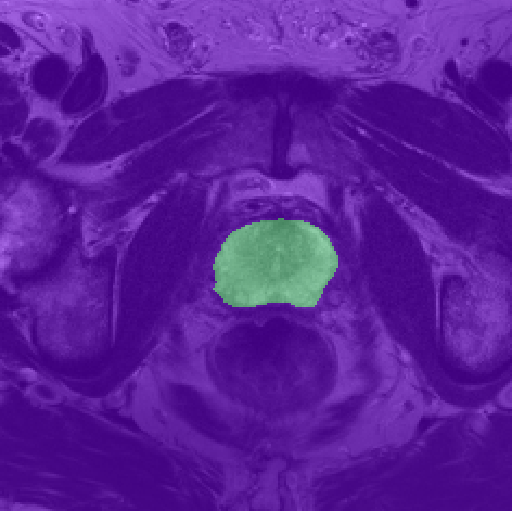}
                        \end{subfigure}
                        \begin{subfigure}[t]{0.19\linewidth}
                                \centering
                                \includegraphics[width=\textwidth, trim=30 30 30 30, clip=true]{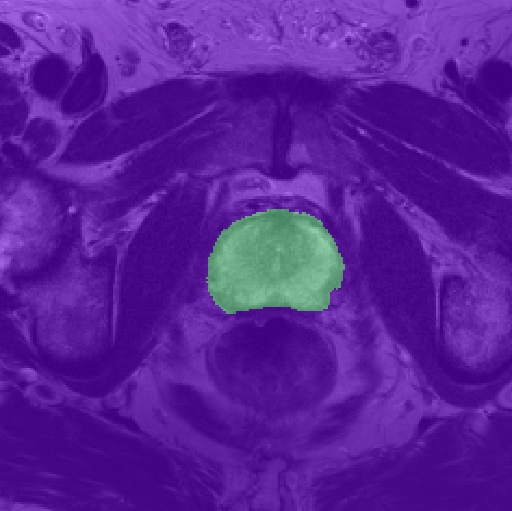}
                        \end{subfigure}
                        \begin{subfigure}[t]{0.19\linewidth}
                                \centering
                                Not applicable
                        \end{subfigure}
                        \\
                        \begin{subfigure}[t]{0.19\linewidth}
                                \centering
                                \includegraphics[width=\textwidth, trim=30 30 30 30, clip=true]{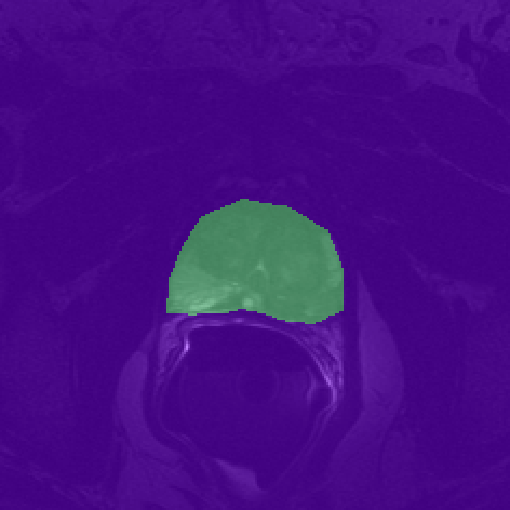}
                                \caption{$y$}
                        \end{subfigure}
                        \begin{subfigure}[t]{0.19\linewidth}
                                \centering
                                \includegraphics[width=\textwidth, trim=30 30 30 30, clip=true]{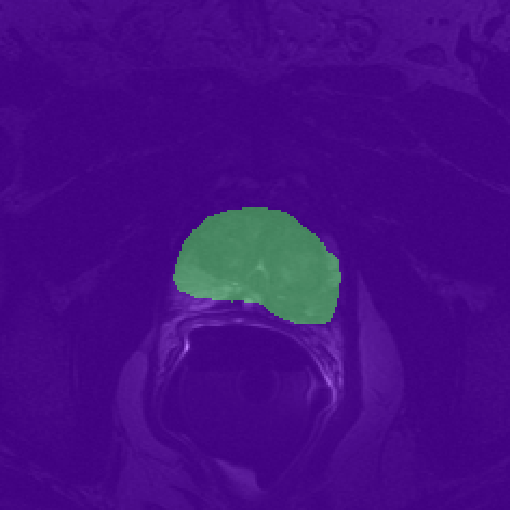}
                                \caption{$\mathcal L_\text{CE}$}
                        \end{subfigure}
                        \begin{subfigure}[t]{0.19\linewidth}
                                \centering
                                \includegraphics[width=\textwidth, trim=30 30 30 30, clip=true]{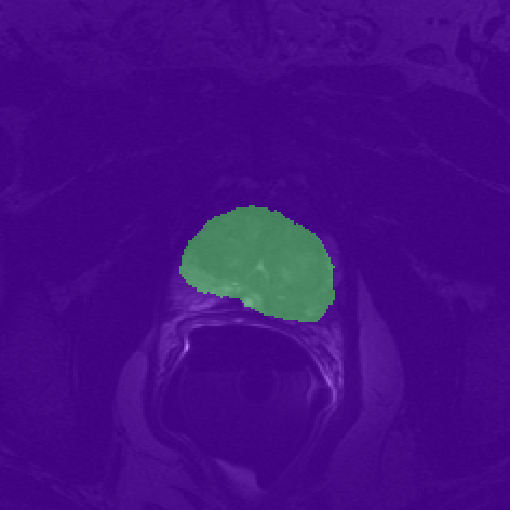}
                                \caption{$\mathcal L_\text{DSC}$}
                        \end{subfigure}
                        \begin{subfigure}[t]{0.19\linewidth}
                                \centering
                                \includegraphics[width=\textwidth, trim=30 30 30 30, clip=true]{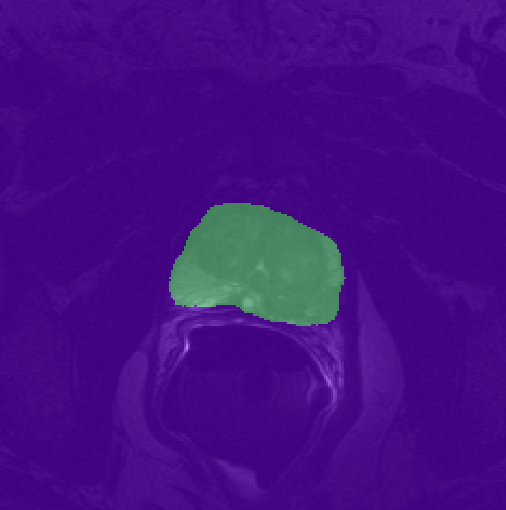}
                                \caption{$\mathcal L_\text{Mime}$}
                        \end{subfigure}
                        \begin{subfigure}[t]{0.19\linewidth}
                                \centering
                                Not applicable
                                \caption{$\mathcal L_\text{NM}$}
                        \end{subfigure}
                        \caption{Additional visualizations of the testing sets.}
                        \label{fig:visual_additional}
                \end{figure}

\end{document}